# Low-cost Stereovision system (disparity map) for few dollars

Rakhmatulin Ildar[1], Eugene Pomazov[2]

[1]South Ural State University, ildar.o2010@yandex.ru

[2]Researcher, StereoPi project

Code source: https://github.com/Ildaron/OpenCV-stereovision-tuner-for-windows

**Abstract**

The paper presents an analysis of the latest developments in the field of stereo vision in the low-cost segment, both for prototypes and for industrial designs. We described the theory of stereo vision and presented information about cameras and data transfer protocols and their compatibility with various devices. The theory in the field of image processing for stereo vision processes is considered and the calibration process is described in detail. Ultimately, we presented the developed stereo vision system and provided the main points that need to be considered when developing such systems. The final, we presented software for adjusting stereo vision parameters in real-time in the python language in the Windows operating system.

**Task** - To develop a compact stereo vision camera with the ability to recognize the distance of weed fruits at a distance of several meters

**Keywords**: OpenCV stereo-vision; low-cost stereo-vision; do it yourself stereo-vision; stereoscopic binocular vision; binocular vision; do it yourself stereo-vision; practical guide stereo-vision

## Introduction

Computer stereo vision is the process of extracting three-dimensional information from digital images by calculating depth based on the binocular discrepancy between the left and right camera images of an object. In stereo vision, two cameras located on the same horizontal line are displaced relative to each other, which allow you to get one image from two points, this works similarly to the binocular vision of a person. Analysis of these two images allows obtaining disparity map information. A disparity map is an indicator of the difference in the relative position of the same points recorded by two cameras. This map allows us to calculate the difference in horizontal coordinates of the corresponding points of the image, which ultimately will allow us to calculate the distance to the object.

Stereovision found wide application in various industries and therefore many works were published on this topic. Oleari et al. [4] presented the paper where images obtained using two cameras - UVC (USB Video device Class). The final cost of the device - 80 euros. Used cameras Logitech C270 can provide images in SVGA resolution (800x600) at 15 frames per second without fully occupying the USB 2.0 bandwidth. Vázquez-Arellano et al. [6] and Rostam et al. [11] presented an overview of stereo vision cameras. These works give a complete understanding of the situation in the stereo vision market today.

Many papers on this topic use the OpenCV standard library (https://opencv.org/) and its tools. Fan et al. [12] described the process of using the OpenCV library for camera calibration. Pomaska [13] used OpenCV with RaspberryPI to determine the distance to the object. The results are presented in the work rather briefly. Similar work was presented by Ke et al. [14] and Chaohui [15], also the authors of these works presented a GUI to easier determine the position of the object.

Our research is focused on finding objects in the fields and is a continuation of our work [20] where stereo vision is necessary to determine the distance to weeds. A similar task was determined in paper [21], where the authors could accurately detect an object at a meter, the results of the work presented in fig. 1.

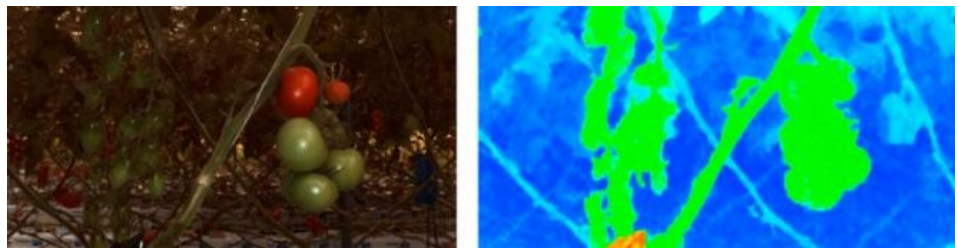

Fig.1. Example of Successful Depth Mapping. Initial image (left) and depth image (right)

But in this work, feature extraction is based on three-dimensional vision, which is very expensive. But the result that was obtained is close to optimal and can serve as a guide for us.

Sarantsev et al. [7] presented a compact stereo camera that captures field scenes and generates three-dimensional point clouds, which are converted into geodetic coordinates and collected into a global map of the field. The paper presents interesting solutions for image processing and in the future, we will try to implement them for our device.

Rovira-Más et al. [8] described a method for creating three-dimensional maps of the terrain by combining information obtained using a stereo camera, a localization sensor, and an inertial measuring unit installed on a platform of mobile equipment. The sensing mechanism contains a compact stereo camera that captures field scenes and generates three-dimensional point clouds, which are converted into geodetic coordinates and collected into a global map of the field. The authors managed to get a good result, but the equipment is too complicated and as a result, the device leaves the low-cost segment.

Kim et al. [9] converted the obtained stereo images into disparity maps by means of stereo matching, and the disparity of each pixel calculated to determine the depth of the distance between the camera and the frame. Depth maps were used to define the edges of regions of interest (ROI), and the clipping regions were segmented using edges located in the expected clipping region closest to the camera, without using any additional markings. Planting heights were calculated using the highest ROI points. This approach was tested on five cultures, and the results showed that the system can detect the target area of the culture even when objects in the resulting images overlap. This paper is close to us in terms of topics, we also use the ROI to isolate the weed and determine the distance to it.

Kise et al. [17] performed field image control by the seamless integration of terrestrial multispectral field images. A vehicle-mounted stereo vision sensor consisting of two multispectral cameras collects images with six spectral channels to form a multispectral three-dimensional field. The work is interesting because the cameras are installed on the vehicle, in real conditions and not laboratory ones.

Ren et al. [18] implemented the collection of images of pests using a PS-15 II lamp and bionic cameras. Based on the detection and comparison of characteristic points in the images of binocular objects, a three-dimensional reconstruction of pest data is achieved. The results presented in the work can also be used for weeds.

Fleischmann et al. [19] in their paper considered such features as the density and distribution of points, as well as the normal of the approximating plane. But despite the positive result, the equipment used in the work is very expensive.

Some works suggest using neural networks instead of stereo vision systems to determine the position of the object, because of which only one camera can be used. But an additional neural network will require an increase in computing power since the main process (GPU) is involved in classification tasks [1]. Also, machine stereo vision is implemented in microcontrollers of the stm32 series and on FPGA - Field-Programmable Gate Array [3].

The main disadvantage of the previously described works is that all the technical aspects are not described in detail. Therefore, it is difficult to repeat experiments to obtain the results claimed by the authors. Any little thing can make a significant impact on the result, so every aspect should be considered. The difference between this manuscript lies in the fact that we have considered and described in detail all the technical aspects of creating a camera. As a result, this manuscript will be useful to researchers who are involved in the development of stereos.

## 2. Implementation
### 2.1 Processor

In view of the task at hand, we need to create a compact and highly efficient device. For our research, we chose the low-cost Jetson Nano single-board computer as the computing processor. The camera can be connected via USB, CSI, or HDMI. The largest selection of cameras for working with Jetson Nano cameras using the CSI protocol is the Camera Serial Interface (CSI), which is largely since this protocol is also used by Raspberry PI, one of the most popular single-board computers. Jetson Nvidia made it physically compatible with Raspberry (15-pin cable with 1mm pitch and 1mm pitch.) CSI is a specification of the Mobile Industry Processor Interface (MIPI). It defines an interface between a camera and a host processor. The problem is that only several cameras with Jetson Nano with preinstalled drivers work with this protocol. Camera drivers can check it on the website:
https://developer.nvidia.com/embedded/jetson-partner-supported-cameras?t1_max-resolution=4K

For example, a fairly old sensor OV2640 binocular camera development board could not be used in projects with jetson nano, fig. 2.

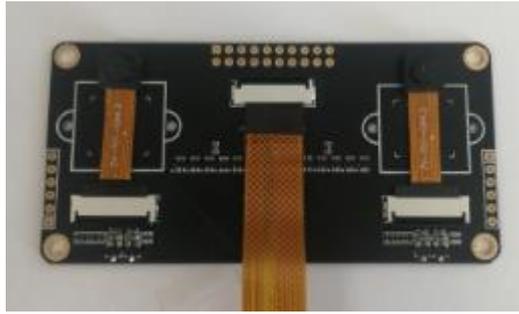

Fig.2. OV2640 binocular camera development board

If there are no drivers for the camera to work via the CSI protocol, there are adapters, for example, a chip - EZ-USB® CX3 Programmable MIPI CSI-2 to USB 3.0 Camera Controller. But to use this chip, you need to prepare a PCB plate. You can also take ready-made boards with a USB adapter, as we did, Fig. 3.

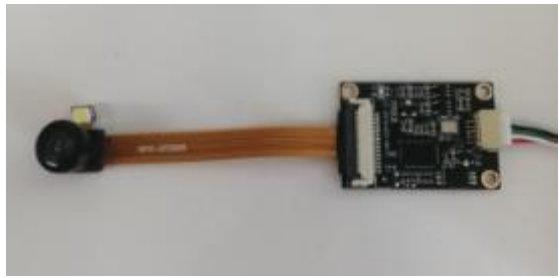

Fig. 3. USB Camera Module 2 Megapixel 60

It should be added that there are many cameras on the market that are implemented for stereo vision tasks. Burdziakowski [2] presented an overview of commercial cameras for stereo vision tasks, only these cameras have a high price.

**2.3 Our implementation**
For this project, we can use picamera, the price of which for the V1 version is only a few dollars. But we made the decision to use CSI cameras connected to the Jetson nano via a USB adapter. Cameras mounted on a sheet of acrylic at the same horizontal line, fig.4.

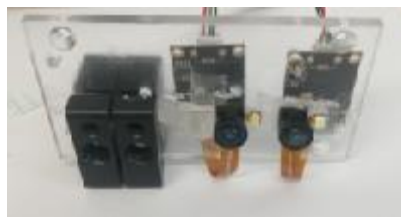

Fig. 4. Developed stereo vision device

The problem with these cameras is that the matrices inside the lens are installed unevenly. Checking whether the cameras are installed correctly, Fig. 5.

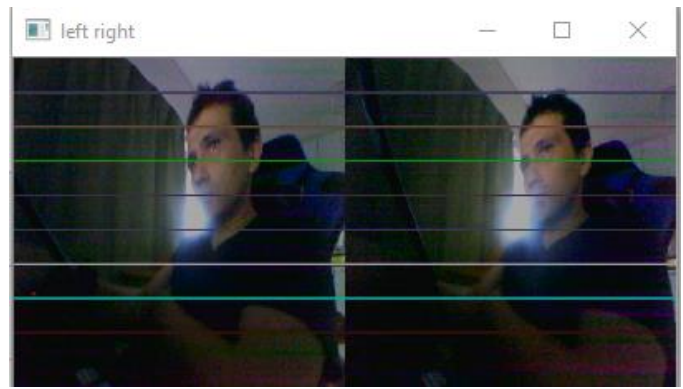

Fig. 5. Checking the correct installation of cameras horizontally

Cameras can be approximately fixed at the same horizontal level, but firmly. Calibration compensates for uneven camera positions.

## 3. Setting
### 3.1. Camera distortion
Ideally, the horizontal and vertical lines in the photo should be parallel throughout the entire image. But in practice, this does not work out and the images are subject to the following distortions:
- Barrel distortion is an optical distortion in which straight lines curve towards the edge of the image;
- Pincushion distortion - this is distortion in which the lines become concave, can be very pronounced at the edges of the picture
- Whisker Distortion - with whisker distortion, horizontal lines protrude up the center and then curve the other way as they get closer to the edge of the frame.
In fig. 6 showed examples of these distortions.

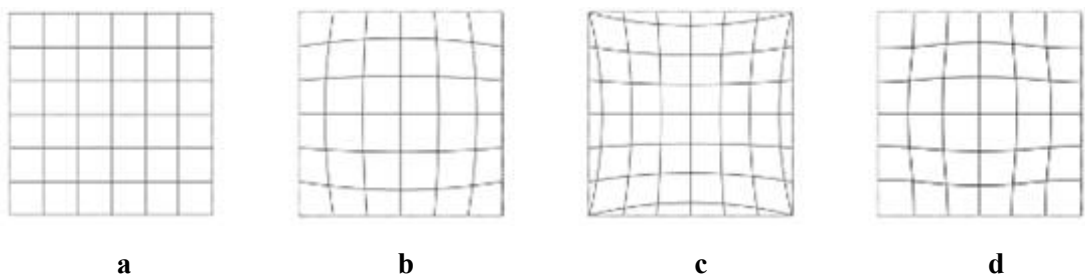

      **a**                      **b**                      **c**                      **d**

Fig. 6. Distortion examples: a - No distortion, b - barell distortion, c - pincushion distortion, d - mustache distortion

### 3.2 Calibrating and configuring a camera
To avoid distortion, the camera must be calibrated. To begin with, we will take 30 photos from each camera from a chessboard measuring 9 by 6. Then, using the function:
retR, cornersR =    cv2.findChessboardCorners cv2.c(outputR,(9,6),None)

- we found the positions of the inner corners of the checkerboard using the sector approach, Fig. 7.

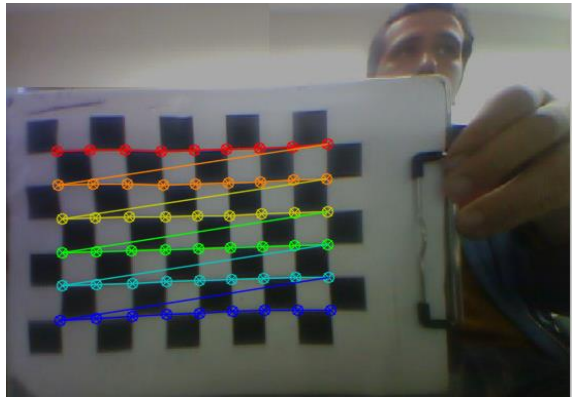

Fig. 7. The process of finding the corners of a checkerboard

We use these corners to calibrate the camera. For calibration, we used the function - cv2.calibrateCamera (). This function returns the camera matrix, distortion factors, rotation, and translation vectors: **retL, mtxL, distL, rvecsL, tvecsL = cv2.calibrateCamera ()**, where:

- distortion coefficients – distL;
- camera matrix – mtxL;
- rotation vectors – rvecsL;
- translation vectors – tvecsL.

We defined the camera matrix based on the free zoom parameter using the following function:

**new_mtxL, roiL = cv2.getOptimalNewCameraMatrix ()**, where:

- new_mtxL– new camera matrix;
- roiL – if need ROI.

### 3.3 Depth map

To compile the depth map, we use the function - cv2. StereoSGBM_create, which based on the SGM algorithm - is a semi-global algorithm that takes global optimization into account when creating a disparity map. Stereo SGBM, consists of the following parameters:

- **minDisparity** - the minimum possible value of the nonconformity;
- **numDisparities -** maximum nonconformity minus minimum nonconformity. This parameter must be divisible by 16;
- **Blocksize -** the appropriate block size. It must be an odd number> = 1;
- **disp12MaxDiff** - the maximum allowable difference (in whole pixels) when checking the discrepancy between the left and right;
- **speckleWindowSize -** the maximum size of mismatch smoothing regions to account for their speckles and invalidate them;
- **speckleRange** - maximum variation of mismatch within each connected component.

## 4. Our software

We created software for a more convenient configuration of the parameters of StereoSGBM in real-time in the windows operating environment. This is software that displays all parameters at once in one window. https://github.com/Ildaron/OpenCV-stereovision-tuner-for-windows. The appearance of the software showed in fig. 8. We have added a pseudo color to the depth map - for the tuning process without it, it is almost impossible to get an adequate visually assessed result.

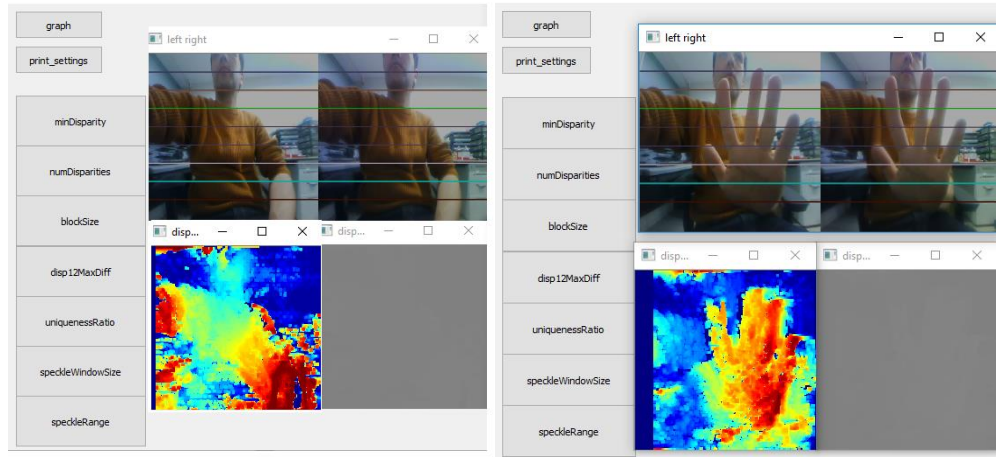

Fig. 8. The appearance of software for adjusting stereo vision parameters (photo with direct light and clean optics)

At the expense of computer vision, the position of the object in the X, Y plane is determined - based on which its ROI area is taken. Then we use stereo vision to compile a depth map and for a given ROI with the NumPy library tool - np.average we calculated the average value for the pixels of this area, which will allow us to calculate the distance to the object.

Depending on the selected camera parameters, the accuracy and sensitivity change depending on the distance of objects from the camera. You can reduce the "visible range" of the camera (for example, from 30 to 80 cm) and increase the accuracy, and vice versa - expand (from 10 cm to 5 meters) but with reduced accuracy. Thus, you can fine-tune the system for specific use.

## 5. Conclusion and suggestions

We provided a complete overview, reviewed protocols, and cameras, and detailed the process of calibrating and configuring cameras. The work showed that it is quite simple to create a stereo vision device. Need only decide which types of cameras, processors, and data transfer protocols you need. After that mechanically install them on any surface in a horizontal equal position and use software that is universal and does not depend on the type of cameras. The cost of the device is $ 30. The cost can be greatly reduced when designing a private PCB board for the camera adapters or use other types of cameras, for example, picameras.

Cameras based on CMOS technologies are installed in many modern devices like camcorders and telephones. These cameras are compact, and it is very easy to install them in conjunction with any devices and use a huge number of settings, about a hundred different registers, to configure the camera, for example, image formats - RAW8 / 10/12/141, YUV422 / 4442, RGB888 / 666/5653, etc. We can use CMOS cameras together with stm32 microcontrollers to create a more reliable, highly efficient, compact, and energy-efficient device. But in this case, the speed of the device will significantly decrease.

**Author Contributions:**

Rakhmatulin, I. - wrote the paper, create device and GUI

Pomazov, E. - technical consultant